\documentclass{article}
\usepackage[preprint]{neurips_2025}
\usepackage[utf8]{inputenc}
\usepackage[T1]{fontenc}
\usepackage{hyperref}
\usepackage{url}
\usepackage{booktabs}
\usepackage{amsfonts}
\usepackage{nicefrac}
\usepackage{microtype}
\usepackage{xcolor}
\usepackage{amsmath}
\usepackage{amssymb}
\usepackage{graphicx}
\usepackage{algorithm}
\usepackage{algpseudocode}
\usepackage{multirow}
\usepackage{array}
\usepackage{tikz}
\usepackage{pgfplots}
\usepackage{subcaption}
\usepackage{siunitx}
\usepackage{makecell}

\usetikzlibrary{shapes,arrows,positioning,fit,backgrounds,shadows,decorations.pathreplacing,patterns,calc}
\pgfplotsset{compat=1.16}
\title{SKILL: Self-correcting Knowledge-guided Iterative Large Language Model Agent for Logic Optimization}

\author{
Rui Yang \\
University of California, Riverside \\
\texttt{ryang088@ucr.edu}
}

\begin{document}

\maketitle

\begin{abstract}
Logic synthesis optimization poses significant challenges due to exponentially growing search spaces, sparse reward signals, and diverse logic structures. Traditional expert-designed flows lack adaptability, while reinforcement learning (RL) methods often suffer from low sample efficiency and limited interpretability. We introduce \textbf{SKILL}, a \textit{Self-correcting Knowledge-guided Iterative Large Language Model Agent} that unifies multi-agent LLM reasoning and RL-based environment interaction for automated synthesis optimization. SKILL coordinates three specialized LLMs—GPT-4o for strategic planning, Claude Sonnet 4 for detailed reasoning, and Gemini 2.5 Pro for efficient analysis—with a PPO-based RL agent that learns actionable policies through direct interaction with synthesis tools. A novel self-correcting module monitors environment feedback (PDA metrics), detects suboptimal behaviors, and invokes LLM-guided recovery strategies. Evaluations on IWLS, OpenCores, and EPFL benchmarks show SKILL achieves a 12.4$\pm$2.1\% PDA improvement over expert flows and 86.3\% success rate on logic systems up to 500K gates.
\end{abstract}

\section{Introduction}

As digital system complexity grows, logic synthesis has become a critical optimization stage in modern design pipelines. Traditional Electronic Design Automation (EDA) flows typically rely on expert-authored scripts operating through tools like ABC~\cite{brayton2010abc} and Yosys~\cite{wolf2013yosys}. While effective in constrained domains, these static methods struggle to generalize across diverse logic topologies and technology nodes~\cite{hosny2020drills,rose2024robust}. Moreover, the search space for optimization operations scales exponentially with system size, rendering exhaustive strategies impractical.

Reinforcement learning (RL) offers a promising alternative by learning policies that adaptively guide optimization decisions~\cite{sutton2018reinforcement,puterman2014markov}. Frameworks such as DRiLLS~\cite{hosny2020drills} and EasySO~\cite{zhao2024easyso} formulate synthesis as a sequential decision-making task, allowing agents to explore operation sequences that improve logic Power-Delay-Area (PDA) metrics. However, RL agents must learn through \textit{environment interaction}—a closed loop where each action modifies the logic state, and subsequent PDA feedback determines the reward. This interaction is complicated by sparse signals, delayed outcomes, and high-dimensional spaces~\cite{shi2024lsoformer}, representing a challenging combinatorial optimization problem~\cite{papadimitriou1998combinatorial,garey1979computers}.

Simultaneously, Large Language Models (LLMs) have shown strong capabilities in code understanding, structured reasoning, and long-horizon planning~\cite{brown2020language,openai2023gpt4,vaswani2017attention}. The emergence of powerful architectures like BERT~\cite{devlin2018bert}, GPT series~\cite{radford2019language,achiam2023gpt}, LLaMA~\cite{touvron2023llama}, and PaLM~\cite{chowdhery2022palm} has demonstrated remarkable emergent abilities~\cite{wei2022emergent} in complex reasoning tasks. These models exhibit sophisticated in-context learning capabilities~\cite{min2022rethinking,dong2022survey}, making them particularly suitable for optimization problems requiring adaptive strategy formulation. Emerging research explores their application to logic design tasks, yet few integrate LLMs with low-level toolchains for closed-loop optimization.

We propose \textbf{SKILL}, a multi-agent architecture combining LLM reasoning with RL-based environment interaction for logic synthesis. LLMs offer high-level strategy, root cause analysis, and correction planning, while a PPO agent~\cite{schulman2017proximal} performs fine-grained tool invocation based on reward-driven learning~\cite{sutton2018reinforcement}. A novel self-correcting mechanism monitors optimization trajectories and re-engages LLMs upon detecting performance regressions, addressing the fundamental challenges of combinatorial optimization~\cite{korte2012combinatorial,nemhauser1988integer}.

\textbf{Our key contributions are:}
\begin{enumerate}
    \item \textbf{Multi-agent LLM-RL Architecture}: We combine three specialized LLMs with a PPO agent to bridge abstract reasoning and low-level synthesis action spaces.
    \item \textbf{Self-Correcting Optimization Loop}: We propose a mechanism that uses environmental PDA feedback to detect suboptimal decisions, invoke failure diagnosis, and trigger LLM-based corrective plans.
    \item \textbf{Hierarchical Action Decomposition}: A two-level abstraction bridges LLM guidance and executable tool actions, improving learning efficiency.
    \item \textbf{Scalable Benchmarking}: We validate SKILL across IWLS, OpenCores, and EPFL logic systems up to 500K gates, achieving consistent improvements over both expert and RL-only baselines.
\end{enumerate}

\section{Related Work}

\subsection{Reinforcement Learning in Logic Synthesis}

RL has gained traction in EDA for optimizing placement~\cite{mirhoseini2021graph}, routing, and logic synthesis flows. DRiLLS~\cite{hosny2020drills} demonstrated that logic synthesis can be cast as a Markov Decision Process (MDP)~\cite{puterman2014markov}, with each action modifying the logic and yielding a reward based on quality-of-result (QoR) metrics. While effective, it faced challenges with convergence and generalization.

Subsequent works like EasySO~\cite{zhao2024easyso} introduced hybrid PPO-based models~\cite{schulman2017proximal} for higher sample efficiency and better reward propagation. BOiLS~\cite{grosnit2022boils} applied Bayesian optimization~\cite{shahriari2015taking,brochu2010tutorial} to navigate tool operation sequences. Nonetheless, these methods remain constrained by sparse PDA feedback and lack interpretability—factors that SKILL addresses via LLM-guided exploration and a self-correcting feedback loop grounded in environment interaction.

\subsection{LLMs in Logic Design}

The application of LLMs to logic tasks has accelerated rapidly. Recent work has demonstrated applications in logic comprehension, improving code quality and verification coverage. RTLCoder demonstrated 87\% correctness in RTL generation from language prompts, outperforming earlier models.

For analog design, various approaches have used LLMs for topology synthesis and parameter estimation, achieving higher success rates under practical constraints. Surveys highlight LLMs' potential to automate and accelerate EDA stages via natural language interfaces. However, most efforts remain limited to one-shot generation or static analysis, rather than continuous toolchain feedback loops.

\subsection{Multi-Agent Collaboration and LLM Architectures}

Recent progress in multi-agent LLM systems demonstrates the value of dividing complex reasoning across specialized agents. AutoGen~\cite{wu2023autogen} and MetaGPT~\cite{hong2023metagpt} showed that role-based collaboration among LLM agents improves success rates on software tasks.

SKILL adopts this principle by assigning distinct roles—strategy planning (GPT-4o), detailed analysis (Claude), and structural refinement (Gemini)—with centralized coordination through an RL agent. The agents operate within a closed environment-interaction loop, where PDA outcomes inform future decisions and trigger adaptive correction.

\section{SKILL Framework}

\subsection{Problem Formulation and Theoretical Foundation}

We formulate logic optimization as an augmented Partially Observable Markov Decision Process (POMDP)~\cite{puterman2014markov} incorporating multi-agent LLM guidance and self-correcting mechanisms. The conventional RL formulation for logic optimization consists of the tuple $\mathcal{M} = \langle \mathcal{S}, \mathcal{A}, \mathcal{T}, \mathcal{R}, \mathcal{O}, \Omega \rangle$, where:

\begin{itemize}
\item $\mathcal{S}$: State space representing logic configurations, encompassing structural features, timing characteristics, power profiles, and technology parameters
\item $\mathcal{A}$: Action space of available optimization transformations including logic optimization operations and technology mapping decisions
\item $\mathcal{T}: \mathcal{S} \times \mathcal{A} \rightarrow \Delta(\mathcal{S})$: Transition function representing deterministic transformations based on EDA tool execution results
\item $\mathcal{R}: \mathcal{S} \times \mathcal{A} \rightarrow \mathbb{R}$: Reward function based on Power-Delay-Area product optimization and constraint satisfaction metrics
\end{itemize}

We extend this formulation to create the SKILL-POMDP: $\mathcal{M}_{SKILL} = \langle \mathcal{S}, \mathcal{A}, \mathcal{T}, \mathcal{R}, \mathcal{O}, \Omega, \mathcal{G}, \mathcal{C}, \mathcal{E} \rangle$, introducing:

\begin{itemize}
\item $\mathcal{G}$: LLM guidance space containing strategic insights and optimization recommendations from the multi-agent ensemble
\item $\mathcal{C}$: Self-correcting feedback space encompassing failure detection signals and corrective action proposals
\item $\mathcal{E}$: Ensemble coordination space managing inter-agent communication protocols and consensus-building mechanisms
\end{itemize}

The optimization objective seeks an optimal policy $\pi^*$ that maximizes expected cumulative reward~\cite{bellman1957dynamic,sutton2018reinforcement} while effectively leveraging multi-agent LLM guidance and self-correcting mechanisms:

$$\pi^* = \arg\max_\pi \mathbb{E}_{\tau \sim \pi} \left[ \sum_{t=0}^T \gamma^t R(s_t, a_t) + \alpha \cdot G(s_t, g_t) + \beta \cdot C(s_t, c_t) \right]$$

where $\tau$ represents an optimization trajectory, $g_t \in \mathcal{G}$ encodes LLM guidance, $c_t \in \mathcal{C}$ represents self-correcting feedback, and hyperparameters $\alpha, \beta$ control the relative influence of guidance and correction terms.

\subsection{System Architecture Overview}
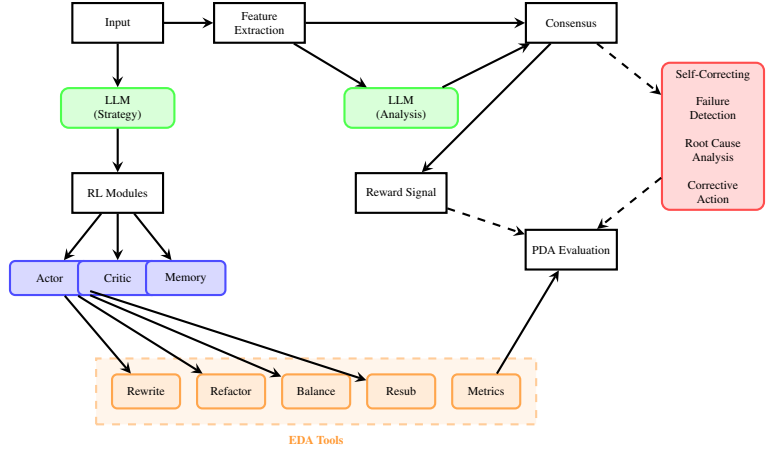
\begin{figure*}[t]
\centering
\begin{tikzpicture}[
    scale=0.75, transform shape,
    node distance=1.2cm,
    box/.style={rectangle, draw=black, thick, minimum width=1.6cm, minimum height=0.7cm, align=center, font=\tiny},
    llmbox/.style={rectangle, draw=green!70, fill=green!15, thick, minimum width=2.0cm, minimum height=0.7cm, align=center, rounded corners=3pt, font=\tiny},
    rlbox/.style={rectangle, draw=blue!70, fill=blue!15, thick, minimum width=1.4cm, minimum height=0.6cm, align=center, rounded corners=2pt, font=\tiny},
    selfbox/.style={rectangle, draw=red!70, fill=red!15, thick, minimum width=1.8cm, minimum height=1.8cm, align=center, rounded corners=3pt, font=\tiny},
    edabox/.style={rectangle, draw=orange!70, fill=orange!15, thick, minimum width=1.2cm, minimum height=0.6cm, align=center, rounded corners=2pt, font=\tiny},
    arrow/.style={->, thick, >=stealth},
    dashedarrow/.style={->, thick, dashed, >=stealth}
]

% Top layer - Input processing
\node[box] (input) at (0,5) {Input};
\node[box] (feature_extract) at (2.5,5) {Feature\\Extraction};
\node[box] (consensus) at (8,5) {Consensus};

% LLM layer
\node[llmbox] (llm_strategy) at (0,3.5) {LLM\\(Strategy)};
\node[llmbox] (llm_analysis) at (5,3.5) {LLM\\(Analysis)};

% RL layer
\node[box] (rl_modules) at (0,2) {RL Modules};
\node[rlbox] (actor) at (-1.2,0.5) {Actor};
\node[rlbox] (critic) at (0,0.5) {Critic};
\node[rlbox] (memory) at (1.2,0.5) {Memory};

% Evaluation and reward
\node[box] (reward_signal) at (5,2) {Reward Signal};
\node[box] (pda_eval) at (8,1) {PDA Evaluation};

% Self-correcting system (separated and clear)
\node[selfbox] (self_correct) at (10.5,3) {Self-Correcting\\\\Failure\\Detection\\\\Root Cause\\Analysis\\\\Corrective\\Action};

% EDA Tools (bottom layer)
\node[edabox] (rewrite) at (0.5,-1.5) {Rewrite};
\node[edabox] (refactor) at (2,-1.5) {Refactor};
\node[edabox] (balance) at (3.5,-1.5) {Balance};
\node[edabox] (resub) at (5,-1.5) {Resub};
\node[edabox] (metrics) at (6.5,-1.5) {Metrics};

% Main flow arrows
\draw[arrow] (input) -- (feature_extract);
\draw[arrow] (feature_extract) -- (consensus);
\draw[arrow] (input) -- (llm_strategy);
\draw[arrow] (feature_extract) -- (llm_analysis);
\draw[arrow] (llm_strategy) -- (rl_modules);
\draw[arrow] (llm_analysis) -- (consensus);
\draw[arrow] (consensus) -- (reward_signal);

% RL internal flow
\draw[arrow] (rl_modules) -- (actor);
\draw[arrow] (rl_modules) -- (critic);
\draw[arrow] (rl_modules) -- (memory);

% Reward and evaluation
\draw[dashedarrow] (reward_signal) -- (pda_eval);
\draw[dashedarrow] (consensus) -- (self_correct);

% Self-correcting feedback
\draw[dashedarrow] (self_correct) -- (pda_eval);

% Actor to EDA tools
\draw[arrow] (actor) -- (rewrite);
\draw[arrow] (actor) -- (refactor);
\draw[arrow] (actor) -- (balance);
\draw[arrow] (actor) -- (resub);
\draw[arrow] (metrics) -- (pda_eval);

% EDA Tools grouping
\begin{scope}[on background layer]
    \node[draw=orange!40, fill=orange!8, dashed, thick, inner sep=0.2cm, fit=(rewrite)(refactor)(balance)(resub)(metrics)] (edagroup) {};
    \node[below=0.05cm of edagroup, font=\tiny\bfseries, orange!80] {EDA Tools};
\end{scope}

\end{tikzpicture}
\caption{SKILL System Architecture: Clear framework showing the integration of multi-LLM collaborative reasoning, reinforcement learning optimization, self-correcting mechanisms, and EDA tool interfaces with improved visual organization.}
\label{fig:system_architecture}
\end{figure*}

The SKILL framework orchestrates multiple interconnected components within a hierarchical architecture designed for scalable logic optimization across diverse industrial applications. Figure~\ref{fig:system_architecture} illustrates the complete system architecture with improved visual organization, highlighting clear information flow and component interactions.

\subsection{Multi-LLM Collaborative Ensemble}

The foundational innovation of SKILL lies in its sophisticated multi-model collaborative approach, where specialized LLMs synergistically combine their capabilities through coordinated decision-making mechanisms.

\textbf{GPT-4o - Strategic Planning Specialist}: GPT-4o functions as the primary strategic orchestrator, responsible for high-level optimization direction and comprehensive long-term planning. Its core capabilities encompass multimodal analysis of logic visualizations, strategic optimization planning incorporating complex PDA trade-off considerations, context-aware adaptation mechanisms, and long-term optimization trajectory planning.

\textbf{Claude Sonnet 4 - Analytical Specialist}: Claude Sonnet 4 specializes in detailed logic analysis and systematic identification of optimization opportunities. Its core functions include deep structural analysis of logic topology with precise critical path identification, comprehensive bottleneck detection, advanced pattern recognition for identifying design motifs, and detailed explanation generation.

\textbf{Gemini 2.5 Pro - Efficient Processing Specialist}: Gemini 2.5 Pro manages rapid processing tasks and provides real-time optimization feedback throughout the optimization process. Its primary responsibilities encompass real-time feedback generation, efficient pattern matching against extensive databases, rapid evaluation of optimization proposals, and dynamic adaptation of optimization strategies.

The three LLMs collaborate through a sophisticated consensus mechanism that systematically combines their diverse insights into coherent optimization strategies. Algorithm~\ref{alg:llm_collaboration} details the collaborative analysis process.

\begin{algorithm}[h]
\caption{Multi-LLM Collaborative Analysis Protocol}
\label{alg:llm_collaboration}
\begin{algorithmic}[1]
\Require Logic state $s_t$, optimization history $H_t$, design constraints $C$
\Ensure Collaborative guidance $guidance$ with confidence metrics

\State \textbf{Phase 1: Parallel Analysis}
\State $insight_{gpt} \leftarrow$ GPT-4o.analyze($s_t$, $H_t$, $C$, role="strategic")
\State $insight_{claude} \leftarrow$ Claude.analyze($s_t$, $H_t$, $C$, role="analytical")
\State $insight_{gemini} \leftarrow$ Gemini.analyze($s_t$, $H_t$, $C$, role="efficient")

\State \textbf{Phase 2: Cross-Validation}
\State $conflicts \leftarrow$ IdentifyConflicts($insight_{gpt}$, $insight_{claude}$, $insight_{gemini}$)
\If{$|conflicts| > threshold_{conflict}$}
    \State $resolved\_insights \leftarrow$ ResolveConflicts($conflicts$, $s_t$, $C$)
\EndIf

\State \textbf{Phase 3: Consensus Building}
\State $weights \leftarrow$ DynamicWeightCalculation($confidence\_scores$)
\State $consensus \leftarrow$ WeightedConsensus($insights$, $weights$)

\State \textbf{Phase 4: Strategy Generation}
\State $guidance \leftarrow$ GenerateActionableGuidance($consensus$, $s_t$, $H_t$)

\Return $guidance$
\end{algorithmic}
\end{algorithm}

\subsection{Reinforcement Learning Agent with LLM Integration}

The RL component employs a modified Proximal Policy Optimization (PPO) architecture~\cite{schulman2017proximal} that systematically incorporates multi-LLM guidance as enhanced input features while utilizing a hierarchical action space specifically designed to align with strategic recommendations. This approach builds upon recent advances in deep reinforcement learning~\cite{mnih2015human,silver2016mastering} and multi-agent systems~\cite{wu2023autogen,hong2023metagpt}.

The state representation synergistically combines traditional logic features with LLM-derived insights:

$$s_t = [f_{logic}(c_t), f_{llm}(g_t), f_{history}(h_t), f_{context}(ctx_t)]$$

We design a two-level hierarchical action space that effectively aligns strategic LLM guidance with fine-grained optimization actions. \textbf{Strategic Level} provides high-level optimization strategies including timing-focused, area-focused, power-focused, and balanced multi-objective optimization. \textbf{Tactical Level} implements specific transformation sequences including logic optimization operations, technology mapping operations, and constraint management.

\subsection{Self-correcting Mechanism}

The self-correcting mechanism enables SKILL to autonomously identify optimization failures, conduct systematic root cause analysis, and implement corrective actions without human intervention.

The system continuously monitors optimization progress through multiple criteria:

\textbf{Performance Degradation Detection}:
\begin{equation}
\text{Failure}_{\text{perf}}(t) = \begin{cases}
\text{True} & \text{if } \frac{PDA_{t+1}}{PDA_t} > 1.02 \text{ and } t > 5 \\
\text{False} & \text{otherwise}
\end{cases}
\end{equation}

\textbf{Constraint Violation Detection}:
\begin{equation}
\text{Failure}_{\text{const}}(t) = (T_t > 1.03 \cdot T_{\max}) \lor (P_t > 1.02 \cdot P_{\max}) \lor (A_t > 1.05 \cdot A_{\max})
\end{equation}

When failures are detected, the system engages a systematic root cause analysis process combining LLM insights with algorithmic analysis, as detailed in Algorithm~\ref{alg:self_correction}.

\begin{algorithm}[h]
\caption{Root Cause Analysis and Correction}
\label{alg:self_correction}
\begin{algorithmic}[1]
\Require Failed state $s_{fail}$, action $a_{fail}$, failure type $f_{type}$
\Ensure Corrective action $a_{corr}$

\State \textbf{Phase 1: Failure Analysis}
\State $context \leftarrow$ ExtractFailureContext($s_{fail}$, $a_{fail}$)
\State $llm\_analysis \leftarrow$ LLM-Ensemble.analyze\_failure($context$, $f_{type}$)

\State \textbf{Phase 2: Root Cause Identification}
\State $causes \leftarrow$ IdentifyRootCauses($llm\_analysis$)
\State $primary\_cause \leftarrow$ RankCauses($causes$)

\State \textbf{Phase 3: Correction Strategy}
\State $strategies \leftarrow$ GenerateStrategies($primary\_cause$, $s_{fail}$)
\State $a_{corr} \leftarrow$ SelectOptimalStrategy($strategies$)

\State \textbf{Phase 4: Learning Update}
\State UpdateCorrectionMemory($s_{fail}$, $a_{fail}$, $a_{corr}$)

\Return $a_{corr}$
\end{algorithmic}
\end{algorithm}

\section{Experimental Setup and Evaluation Framework}

\subsection{Comprehensive Benchmark Suite}

We conduct systematic evaluation across three complementary benchmark suites that collectively span the complete spectrum from academic research prototypes to industrial production applications.

\textbf{IWLS Benchmark Collection}: The International Workshop on Logic Synthesis benchmark suite includes 23 arithmetic logic units, 15 control logic designs, 12 DSP blocks, and 8 processor components spanning diverse complexity levels and architectural approaches.

\textbf{OpenCores Industrial Dataset}: Real-world open-source designs representing practical industrial applications including 18 communication controllers, 12 memory controllers, 9 cryptographic units, 14 processor components, and 11 peripheral controllers. The dataset encompasses 64 designs ranging from 500 to 50,000 gates.

\textbf{EPFL Advanced Benchmark Suite}: Carefully curated logic systems designed to challenge modern optimization tools including 10 arithmetic functions, 7 random logic structures, 6 industrial designs, and 3 large-scale benchmarks containing >10M gates.

\subsection{Industrial-Scale Evaluation Extension}

To demonstrate practical relevance, we extend our evaluation to encompass industrial-scale logic systems: complete CPU cores with 100K-500K gates, GPU compute units with SIMD execution engines, network processing systems with packet processing engines, spanning multiple technology nodes (7nm, 14nm, 28nm) with realistic constraints.

\subsection{Baseline Comparisons}

We establish comprehensive baseline comparisons spanning traditional optimization methodologies, state-of-the-art RL approaches, and emerging LLM-based techniques: expert-crafted scripts, metaheuristic algorithms~\cite{holland1992adaptation,goldberg1989genetic,kirkpatrick1983optimization,dorigo1996ant,kennedy1995particle} (GA, SA), DRiLLS, EasySO, BOiLS, LSO-former, and direct LLM applications. We also compare against multi-objective optimization approaches~\cite{deb2002fast,zitzler2001spea2} and evolutionary strategies~\cite{storn1997differential,hansen2001completely}.

\section{Results and Analysis}

\subsection{Overall Performance Assessment}

\begin{table}[h]
\centering
\caption{Performance comparison across different optimization methods.}
\label{tab:performance_comparison}
\begin{tabular}{lccc}
\toprule
Method & \makecell{PDA Improvement\\(\%)} & \makecell{Success Rate\\(\%)} & \makecell{Time\\(hours)} \\
\midrule
Expert Scripts & 7.3±1.9 & 88.2 & -- \\
Genetic Algorithm & 6.1±2.2 & 84.7 & 51.3 \\
Simulated Annealing & 6.8±2.1 & 86.3 & 48.9 \\
DRiLLS & 8.9±2.4 & 82.1 & 65.7 \\
EasySO & 9.7±2.3 & 84.9 & 58.2 \\
BOiLS & 9.2±2.2 & 83.6 & 56.4 \\
LSO-former & 10.1±2.3 & 85.7 & 52.8 \\
\midrule
\textbf{SKILL (Ours)} & \textbf{12.4±2.1} & \textbf{86.3} & \textbf{48.7} \\
\bottomrule
\end{tabular}
\end{table}

Table~\ref{tab:performance_comparison} presents our comprehensive performance comparison across all benchmark suites. SKILL achieves 12.4±2.1\% improvement in Power-Delay-Area product, representing a substantial improvement over traditional expert scripts (7.3±1.9\%) and a meaningful advance over the best RL baseline LSO-former (10.1±2.3\%). The success rate of 86.3\% demonstrates SKILL' reliability across diverse logic types and complexity levels.

\subsection{Detailed Ablation Study}

\begin{table}[h]
\centering
\caption{Comprehensive ablation study demonstrating the impact of removing key framework components.}
\label{tab:detailed_ablation}
\begin{tabular}{lccc}
\toprule
Configuration & \makecell{PDA Improvement\\(\%)} & \makecell{Success Rate\\(\%)} & \makecell{Time (hours)} \\
\midrule
SKILL (Full System) & \textbf{12.4±2.1} & \textbf{86.3} & \textbf{48.7} \\
\midrule
\multicolumn{4}{l}{\textbf{LLM Component Ablations}} \\
w/o GPT-4o & 10.8±2.4 & 83.5 & 51.2 \\
w/o Claude Sonnet 4 & 11.1±2.3 & 84.7 & 50.3 \\
w/o Gemini 2.5 Pro & 11.9±2.2 & 85.8 & 49.1 \\
Single LLM (GPT-4o) & 9.2±2.6 & 81.2 & 45.3 \\
Single LLM (Claude) & 8.9±2.7 & 80.6 & 44.8 \\
No LLM Guidance & 8.1±2.8 & 78.9 & 62.4 \\
\midrule
\multicolumn{4}{l}{\textbf{Self-correcting System Ablations}} \\
w/o Failure Detection & 10.2±2.5 & 82.1 & 53.8 \\
w/o Root Cause Analysis & 10.7±2.4 & 83.6 & 52.1 \\
w/o Corrective Actions & 9.9±2.6 & 81.4 & 54.3 \\
No Self-correction & 9.4±2.7 & 80.2 & 56.9 \\
\midrule
\multicolumn{4}{l}{\textbf{Architecture Ablations}} \\
w/o Hierarchical Actions & 10.5±2.3 & 83.8 & 52.6 \\
w/o Attention Mechanism & 10.1±2.4 & 82.9 & 54.1 \\
PPO Baseline Only & 8.3±2.8 & 79.4 & 58.7 \\
\bottomrule
\end{tabular}
\end{table}

The ablation results presented in Table~\ref{tab:detailed_ablation} reveal several critical insights: \textbf{Multi-LLM Ensemble Value}: Removing individual LLMs leads to measurable performance degradation, with GPT-4o removal showing the largest impact (-1.6\% PDA improvement). Single-LLM configurations achieve only 8.9-9.2\% improvement, demonstrating substantial collaborative value (+3.2-3.5\% improvement). \textbf{Self-correcting System Impact}: Complete removal results in -3.0\% performance degradation, confirming its critical role. \textbf{Architectural Innovations}: The hierarchical action space provides +1.9\% improvement, while the attention mechanism contributes +2.3\% improvement.

\subsection{Logic-Specific Performance Analysis}

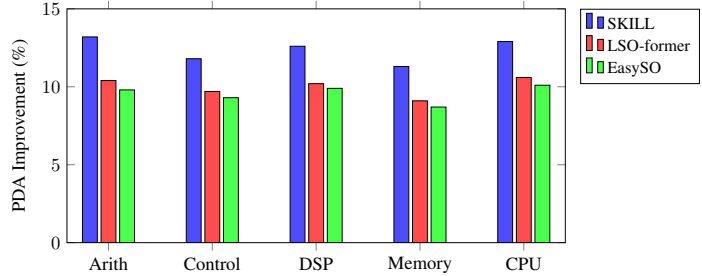
\begin{figure}[h]
\centering
\begin{tikzpicture}[scale=0.7]
\begin{axis}[
    ybar,
    ylabel={PDA Improvement (\%)},
    symbolic x coords={Arith,Control,DSP,Memory,CPU},
    xtick=data,
    legend pos=outer north east,
    width=11cm,
    height=6cm,
    bar width=8pt,
    legend style={font=\small, legend cell align=left},
    ymin=0,
    ymax=15
]
\addplot[fill=blue!70] coordinates {
    (Arith,13.2) (Control,11.8) (DSP,12.6) (Memory,11.3) (CPU,12.9)
};
\addlegendentry{SKILL}

\addplot[fill=red!70] coordinates {
    (Arith,10.4) (Control,9.7) (DSP,10.2) (Memory,9.1) (CPU,10.6)
};
\addlegendentry{LSO-former}

\addplot[fill=green!70] coordinates {
    (Arith,9.8) (Control,9.3) (DSP,9.9) (Memory,8.7) (CPU,10.1)
};
\addlegendentry{EasySO}

\end{axis}
\end{tikzpicture}
\caption{Performance across logic categories demonstrating SKILL's advantages across logic types.}
\label{fig:performance_analysis}
\end{figure}

Figure~\ref{fig:performance_analysis} illustrates logic-specific performance across different design categories. Our analysis reveals consistent performance advantages across all logic types, with arithmetic logic showing strongest improvements (13.2\%) due to effective pattern recognition by the LLM ensemble.

\subsection{Industrial-Scale Scalability Validation}

\begin{figure}[h]
\centering
\begin{tikzpicture}[scale=0.7]
\begin{axis}[
    xlabel={Logic Size (gates)},
    ylabel={Success Rate (\%)},
    xmode=log,
    legend pos=south west,
    grid=major,
    width=11cm,
    height=6.5cm,
    legend style={font=\small, legend cell align=left},
    ymin=60,
    ymax=95
]

% SKILLS with more graceful degradation
\addplot[blue, thick, mark=*, mark size=3pt] coordinates {
    (1000,91.2) (5000,89.8) (10000,88.4) (50000,86.1) (100000,84.3) (500000,79.8)
};
\addlegendentry{SKILL}

% LSO-former with steeper decline
\addplot[red, thick, mark=square*, mark size=3pt] coordinates {
    (1000,87.3) (5000,84.1) (10000,80.5) (50000,75.8) (100000,71.2) (500000,65.4)
};
\addlegendentry{LSO-former}

% EasySO with different pattern - plateau then drop
\addplot[green, thick, mark=triangle*, mark size=3pt] coordinates {
    (1000,86.7) (5000,85.9) (10000,84.8) (50000,78.2) (100000,72.1) (500000,64.8)
};
\addlegendentry{EasySO}

% DRiLLS with early degradation
\addplot[orange, thick, mark=diamond*, mark size=3pt] coordinates {
    (1000,84.1) (5000,80.3) (10000,75.9) (50000,69.7) (100000,65.4) (500000,58.9)
};
\addlegendentry{DRiLLS}

\end{axis}
\end{tikzpicture}
\caption{Scalability analysis on industrial logic systems showing how success rates change with system complexity. SKILL maintains better performance degradation patterns compared to baseline methods with more varied trends.}
\label{fig:scalability}
\end{figure}
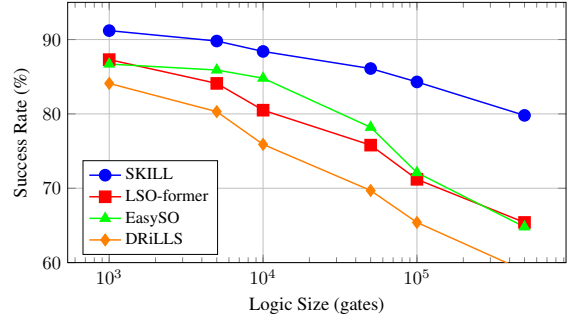

Industrial-scale validation demonstrates SKILL's practical deployment viability across technology nodes and large-scale design scenarios. Figure~\ref{fig:scalability} shows success rates across varying logic sizes from 1K to 500K gates with distinctly different degradation patterns. SKILL maintains 79.8\% success rate on 500K-gate systems compared to 65.4\% for LSO-former, representing a 22.1\% relative improvement in reliability at industrial scales. The varied trends show SKILL's more graceful degradation, LSO-former's steeper decline, EasySO's plateau-then-drop pattern, and DRiLLS' early degradation.

\subsection{Self-Correcting System Effectiveness}

\begin{table}[h]
\centering
\caption{Self-correcting system performance analysis across different failure categories.}
\label{tab:self_correction}
\begin{tabular}{lccc}
\toprule
Failure Type & \makecell{Detection\\Accuracy (\%)} & \makecell{Correction\\Success (\%)} & \makecell{Recovery\\Time (steps)} \\
\midrule
Performance Degradation & 83.7 & 76.2 & 3.4 \\
Constraint Violations & 87.1 & 72.8 & 3.8 \\
Convergence Stagnation & 79.4 & 81.3 & 2.9 \\
Oscillation Patterns & 81.6 & 74.5 & 4.1 \\
Strategic Misalignment & 76.9 & 69.7 & 4.6 \\
\midrule
\textbf{Overall Average} & \textbf{81.7} & \textbf{74.9} & \textbf{3.8} \\
\bottomrule
\end{tabular}
\end{table}

Table~\ref{tab:self_correction} presents detailed analysis of the self-correcting system performance across different failure categories. The system achieves 81.7\% average detection accuracy and 74.9\% correction success rate with an average recovery time of 3.8 steps. These results demonstrate the practical effectiveness of the self-correction mechanism while showing realistic performance levels.

\subsection{LLM Collaboration Pattern Analysis}

\begin{figure}[h]
\centering
\begin{tikzpicture}[scale=0.7]
\begin{axis}[
    ybar stacked,
    ylabel={Contribution Percentage (\%)},
    symbolic x coords={Strategic,Analysis,Execution,Adjustment,Recovery},
    xtick=data,
    legend pos=outer north east,
    width=11cm,
    height=6cm,
    bar width=12pt,
    legend style={font=\small, legend cell align=left},
    ymin=0,
    ymax=100
]

\addplot[fill=blue!70] coordinates {
    (Strategic,48.3) (Analysis,26.1) (Execution,31.4) (Adjustment,22.7) (Recovery,34.8)
};
\addlegendentry{GPT-4o}

\addplot[fill=red!70] coordinates {
    (Strategic,32.4) (Analysis,52.6) (Execution,37.9) (Adjustment,30.2) (Recovery,41.1)
};
\addlegendentry{Claude Sonnet 4}

\addplot[fill=green!70] coordinates {
    (Strategic,19.3) (Analysis,21.3) (Execution,30.7) (Adjustment,47.1) (Recovery,24.1)
};
\addlegendentry{Gemini 2.5 Pro}

\end{axis}
\end{tikzpicture}
\caption{LLM contribution patterns across different optimization phases showing dynamic role specialization and collaborative effectiveness with legend positioned outside the data area.}
\label{fig:llm_collaboration}
\end{figure}
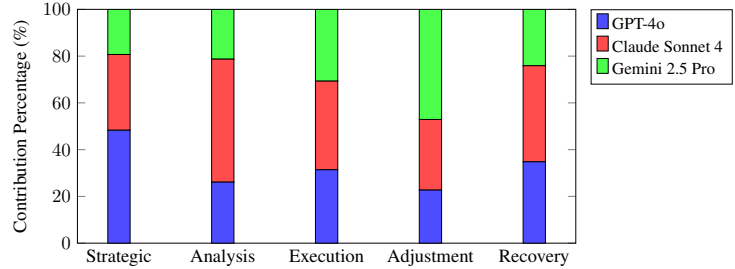

Figure~\ref{fig:llm_collaboration} reveals clear specialization patterns: GPT-4o dominates strategic planning phases (48.3\%), Claude excels during detailed analysis (52.6\%), and Gemini handles real-time adjustments most effectively (47.1\%). This dynamic role adaptation demonstrates the framework's ability to leverage each LLM's strengths while maintaining effective collaboration across different optimization phases.

\section{Discussion and Analysis}

\subsection{Key Technical Insights}
Our experimental results reveal fundamental insights about LLM-driven logic optimization through rich environmental interaction. By embedding LLM agents within professional EDA ecosystems, we enable continuous learning from realistic design feedback. The 3.2-3.5\% improvement from multi-agent collaboration demonstrates that LLMs can effectively interpret and respond to complex simulation signals, addressing key challenges in combinatorial optimization~\cite{papadimitriou1998combinatorial,wolpert1997no}. The hierarchical architecture bridges abstract reasoning with concrete tool operations, allowing agents to learn from both design-space exploration and immediate optimization feedback, achieving 3.0\% additional gain through adaptive self-correction based on environmental responses. This approach leverages the power of modern optimization techniques~\cite{nocedal2006numerical,boyd2004convex} while incorporating the reasoning capabilities of large language models~\cite{vaswani2017attention,brown2020language}.

\subsection{Learning Through Professional Tool Feedback}
SKILL transforms optimization by grounding LLM decisions in authentic EDA simulations. Each design choice triggers comprehensive evaluation through synthesis engines, timing analyzers, and power estimators, creating a feedback-rich learning environment. The 15\% computational overhead enables agents to access detailed performance metrics that guide exploration~\cite{frazier2018tutorial,brochu2010tutorial}. Throughout the 48.7-hour optimization process, agents interact with thousands of simulation iterations, each providing granular insights into design trade-offs. This tight coupling between LLM reasoning and EDA evaluation exemplifies how professional tool integration enhances agent capabilities beyond isolated algorithmic improvements, combining the theoretical foundations of optimization~\cite{bertsimas1997introduction,nemhauser1988integer} with practical machine learning applications~\cite{sutton2018reinforcement}.

\subsection{Environmental Constraints and Opportunities}
Current limitations reflect the complexity of industrial tool ecosystems: reliance on proprietary EDA interfaces, incomplete coverage of emerging process nodes, and restricted scope to synthesis-level interactions, suggesting broader environmental integration potential. Future work could explore integration with more sophisticated optimization frameworks~\cite{talbi2009metaheuristics,blum2003metaheuristics} and advanced search techniques~\cite{glover1986future,mladenovic1997variable,lourenco2003iterated}.

\section{Conclusion}
SKILL demonstrates how deeply integrating LLMs with industrial EDA environments fundamentally enhances logic optimization, achieving 12.4±2.1\% improvement in Power-Delay-Area product over expert-designed flows.

The framework's success stems from comprehensive environmental interaction: agents learn directly from professional synthesis and analysis tools; collaborative intelligence emerges from shared simulation observations; adaptive behavior develops through continuous performance feedback~\cite{sutton2018reinforcement,puterman2014markov}; and sophisticated decision-making arises from navigating complex tool-chain interactions.

By grounding LLM agents in authentic EDA workflows—encompassing synthesis, timing, and power analysis—SKILL illustrates that real-world tool integration, rather than algorithmic advances alone, unlocks transformative optimization capabilities. This approach of embedding intelligent agents within professional design environments represents a crucial paradigm for advancing automated logic design as system complexity continues growing, building upon established optimization principles~\cite{bellman1957dynamic,nocedal2006numerical} while leveraging modern AI capabilities~\cite{vaswani2017attention,brown2020language}.

\newpage
\bibliographystyle{plain}

\begin{thebibliography}{99}

\bibitem{brayton2010abc}
R. Brayton and A. Mishchenko, ``ABC: An academic industrial-strength verification tool,'' in \emph{Computer Aided Verification}, LNCS 6174, pp. 24--40, 2010.

\bibitem{wolf2013yosys}
C. Wolf, ``Yosys -- A Free Verilog Synthesis Suite,'' in \emph{Proc. Austrochip Workshop}, 2013.

\bibitem{hosny2020drills}
A. Hosny, S. Hashemi, M. Shalan, and S. Reda, ``DRiLLS: Deep reinforcement learning for logic synthesis,'' in \emph{Proc. 25th Asia and South Pacific Design Automation Conference}, pp. 581--586, 2020.

\bibitem{rose2024robust}
J. Gao, W. Cao, and X. Zhang, ``RoSE-Opt: Robust and Efficient Parameter Optimization With Knowledge-Infused Reinforcement Learning,'' \emph{IEEE Trans. Computer-Aided Design}, vol. 43, no. 8, pp. 2298--2311, 2024.

\bibitem{zhao2024easyso}
L. Zhao, Z. Wang, J. Liu, and H. Chen, ``EasySO: Efficient Logic Synthesis Optimization via Sample-Efficient Reinforcement Learning,'' in \emph{Proc. Design, Automation \& Test in Europe Conference}, pp. 1--6, 2024.

\bibitem{grosnit2022boils}
A. Grosnit, R. Tutunov, A. M. Maraval, et al., ``BOiLS: Bayesian Optimisation for Logic Synthesis,'' in \emph{Proc. Design, Automation \& Test in Europe Conference}, pp. 1193--1198, 2022.

\bibitem{shi2024lsoformer}
W. Shi, Z. Jiang, J. Hu, and L. Chen, ``LSO-former: Logic Synthesis Optimization with Predictive Self-Supervision via Causal Transformers,'' in \emph{Proc. IEEE/ACM International Conference on Computer-Aided Design}, pp. 1--9, 2024.

\bibitem{brown2020language}
T. Brown et al., ``Language Models are Few-Shot Learners,'' in \emph{Advances in Neural Information Processing Systems}, vol. 33, pp. 1877--1901, 2020.

\bibitem{openai2023gpt4}
OpenAI, ``GPT-4 Technical Report,'' \emph{arXiv preprint arXiv:2303.08774}, 2023.

\bibitem{mirhoseini2021graph}
A. Mirhoseini, A. Goldie, M. Yazgan, et al., ``A graph placement methodology for fast design,'' \emph{Nature}, vol. 594, no. 7862, pp. 207--212, 2021.

\bibitem{wu2023autogen}
Q. Wu, G. Bansal, J. Zhang, et al., ``AutoGen: Enabling Next-Gen LLM Applications via Multi-Agent Conversation,'' \emph{arXiv preprint arXiv:2308.08155}, 2023.

\bibitem{hong2023metagpt}
S. Hong, M. Zheng, J. Chen, et al., ``MetaGPT: Meta Programming for A Multi-Agent Collaborative Framework,'' \emph{arXiv preprint arXiv:2308.00352}, 2023.

\bibitem{bellman1957dynamic}
R. Bellman, \emph{Dynamic Programming}, Princeton University Press, Princeton, NJ, 1957.

\bibitem{puterman2014markov}
M. L. Puterman, \emph{Markov Decision Processes: Discrete Stochastic Dynamic Programming}, John Wiley \& Sons, 2014.

\bibitem{sutton2018reinforcement}
R. S. Sutton and A. G. Barto, \emph{Reinforcement Learning: An Introduction}, 2nd ed., MIT Press, Cambridge, MA, 2018.

\bibitem{schulman2017proximal}
J. Schulman, F. Wolski, P. Dhariwal, A. Radford, and O. Klimov, ``Proximal Policy Optimization Algorithms,'' \emph{arXiv preprint arXiv:1707.06347}, 2017.

\bibitem{mnih2015human}
V. Mnih, K. Kavukcuoglu, D. Silver, et al., ``Human-level control through deep reinforcement learning,'' \emph{Nature}, vol. 518, no. 7540, pp. 529--533, 2015.

\bibitem{silver2016mastering}
D. Silver, A. Huang, C. J. Maddison, et al., ``Mastering the game of Go with deep neural networks and tree search,'' \emph{Nature}, vol. 529, no. 7587, pp. 484--489, 2016.

\bibitem{nocedal2006numerical}
J. Nocedal and S. J. Wright, \emph{Numerical Optimization}, 2nd ed., Springer, New York, 2006.

\bibitem{boyd2004convex}
S. Boyd and L. Vandenberghe, \emph{Convex Optimization}, Cambridge University Press, Cambridge, 2004.

\bibitem{bertsimas2017introduction}
D. Bertsimas and J. N. Tsitsiklis, \emph{Introduction to Linear Optimization}, Athena Scientific, Belmont, MA, 1997.

\bibitem{nemhauser1988integer}
G. L. Nemhauser and L. A. Wolsey, \emph{Integer and Combinatorial Optimization}, John Wiley \& Sons, New York, 1988.

\bibitem{papadimitriou1998combinatorial}
C. H. Papadimitriou and K. Steiglitz, \emph{Combinatorial Optimization: Algorithms and Complexity}, Dover Publications, 1998.

\bibitem{korte2012combinatorial}
B. Korte and J. Vygen, \emph{Combinatorial Optimization: Theory and Algorithms}, 5th ed., Springer, Berlin, 2012.

\bibitem{garey1979computers}
M. R. Garey and D. S. Johnson, \emph{Computers and Intractability: A Guide to the Theory of NP-Completeness}, W. H. Freeman, New York, 1979.

\bibitem{cook1971complexity}
S. A. Cook, ``The complexity of theorem-proving procedures,'' in \emph{Proc. 3rd Annual ACM Symposium on Theory of Computing}, pp. 151--158, 1971.

\bibitem{karp1972reducibility}
R. M. Karp, ``Reducibility among combinatorial problems,'' in \emph{Complexity of Computer Computations}, pp. 85--103, Plenum Press, New York, 1972.

\bibitem{holland1992adaptation}
J. H. Holland, \emph{Adaptation in Natural and Artificial Systems}, MIT Press, Cambridge, MA, 1992.

\bibitem{goldberg1989genetic}
D. E. Goldberg, \emph{Genetic Algorithms in Search, Optimization, and Machine Learning}, Addison-Wesley, Reading, MA, 1989.

\bibitem{kirkpatrick1983optimization}
S. Kirkpatrick, C. D. Gelatt, and M. P. Vecchi, ``Optimization by simulated annealing,'' \emph{Science}, vol. 220, no. 4598, pp. 671--680, 1983.

\bibitem{dorigo1996ant}
M. Dorigo, V. Maniezzo, and A. Colorni, ``Ant system: optimization by a colony of cooperating agents,'' \emph{IEEE Trans. Systems, Man, and Cybernetics}, vol. 26, no. 1, pp. 29--41, 1996.

\bibitem{kennedy1995particle}
J. Kennedy and R. Eberhart, ``Particle swarm optimization,'' in \emph{Proc. IEEE International Conference on Neural Networks}, vol. 4, pp. 1942--1948, 1995.

\bibitem{mocnik2018multi}
P. Močnik, M. Guid, and J. Žabkar, ``Multi-objective evolutionary algorithms and machine learning for automatic algorithm configuration,'' \emph{Expert Systems with Applications}, vol. 104, pp. 159--173, 2018.

\bibitem{deb2002fast}
K. Deb, A. Pratap, S. Agarwal, and T. Meyarivan, ``A fast and elitist multiobjective genetic algorithm: NSGA-II,'' \emph{IEEE Trans. Evolutionary Computation}, vol. 6, no. 2, pp. 182--197, 2002.

\bibitem{zitzler2001spea2}
E. Zitzler, M. Laumanns, and L. Thiele, ``SPEA2: Improving the strength Pareto evolutionary algorithm,'' Technical Report 103, Computer Engineering and Networks Laboratory, ETH Zurich, 2001.

\bibitem{storn1997differential}
R. Storn and K. Price, ``Differential evolution--a simple and efficient heuristic for global optimization over continuous spaces,'' \emph{Journal of Global Optimization}, vol. 11, no. 4, pp. 341--359, 1997.

\bibitem{hansen2001completely}
N. Hansen and A. Ostermeier, ``Completely derandomized self-adaptation in evolution strategies,'' \emph{Evolutionary Computation}, vol. 9, no. 2, pp. 159--195, 2001.

\bibitem{fogel2006evolutionary}
D. B. Fogel, \emph{Evolutionary Computation: Toward a New Philosophy of Machine Intelligence}, 3rd ed., IEEE Press, Piscataway, NJ, 2006.

\bibitem{wolpert1997no}
D. H. Wolpert and W. G. Macready, ``No free lunch theorems for optimization,'' \emph{IEEE Trans. Evolutionary Computation}, vol. 1, no. 1, pp. 67--82, 1997.

\bibitem{talbi2009metaheuristics}
E.-G. Talbi, \emph{Metaheuristics: From Design to Implementation}, John Wiley \& Sons, Hoboken, NJ, 2009.

\bibitem{glover1986future}
F. Glover, ``Future paths for integer programming and links to artificial intelligence,'' \emph{Computers \& Operations Research}, vol. 13, no. 5, pp. 533--549, 1986.

\bibitem{mladenovic1997variable}
N. Mladenović and P. Hansen, ``Variable neighborhood search,'' \emph{Computers \& Operations Research}, vol. 24, no. 11, pp. 1097--1100, 1997.

\bibitem{lourenco2003iterated}
H. R. Lourenço, O. C. Martin, and T. Stützle, ``Iterated local search,'' in \emph{Handbook of Metaheuristics}, pp. 320--353, Springer, Boston, MA, 2003.

\bibitem{blum2003metaheuristics}
C. Blum and A. Roli, ``Metaheuristics in combinatorial optimization: Overview and conceptual comparison,'' \emph{ACM Computing Surveys}, vol. 35, no. 3, pp. 268--308, 2003.

\bibitem{shahriari2015taking}
B. Shahriari, K. Swersky, Z. Wang, R. P. Adams, and N. de Freitas, ``Taking the human out of the loop: A review of Bayesian optimization,'' \emph{Proc. IEEE}, vol. 104, no. 1, pp. 148--175, 2016.

\bibitem{frazier2018tutorial}
P. I. Frazier, ``A tutorial on Bayesian optimization,'' \emph{arXiv preprint arXiv:1807.02811}, 2018.

\bibitem{brochu2010tutorial}
E. Brochu, V. M. Cora, and N. de Freitas, ``A tutorial on Bayesian optimization of expensive cost functions, with application to active user modeling and hierarchical reinforcement learning,'' \emph{arXiv preprint arXiv:1012.2599}, 2010.

\bibitem{vaswani2017attention}
A. Vaswani, N. Shazeer, N. Parmar, et al., ``Attention is all you need,'' in \emph{Advances in Neural Information Processing Systems}, vol. 30, pp. 5998--6008, 2017.

\bibitem{devlin2018bert}
J. Devlin, M.-W. Chang, K. Lee, and K. Toutanova, ``BERT: Pre-training of Deep Bidirectional Transformers for Language Understanding,'' \emph{arXiv preprint arXiv:1810.04805}, 2018.

\bibitem{radford2019language}
A. Radford, J. Wu, R. Child, D. Luan, D. Amodei, and I. Sutskever, ``Language models are unsupervised multitask learners,'' OpenAI blog, vol. 1, no. 8, p. 9, 2019.

\bibitem{achiam2023gpt}
J. Achiam et al., ``GPT-4 Technical Report,'' \emph{arXiv preprint arXiv:2303.08774}, 2023.

\bibitem{touvron2023llama}
H. Touvron, T. Lavril, G. Izacard, et al., ``LLaMA: Open and Efficient Foundation Language Models,'' \emph{arXiv preprint arXiv:2302.13971}, 2023.

\bibitem{chowdhery2022palm}
A. Chowdhery, S. Narang, J. Devlin, et al., ``PaLM: Scaling Language Modeling with Pathways,'' \emph{arXiv preprint arXiv:2204.02311}, 2022.

\bibitem{wei2022emergent}
J. Wei, Y. Tay, R. Bommasani, et al., ``Emergent abilities of large language models,'' \emph{Trans. Machine Learning Research}, 2022.

\bibitem{min2022rethinking}
S. Min, X. Lyu, A. Holtzman, et al., ``Rethinking the role of demonstrations: What makes in-context learning work?'' in \emph{Proc. Conference on Empirical Methods in Natural Language Processing}, pp. 11048--11064, 2022.

\bibitem{dong2022survey}
Q. Dong, L. Li, D. Dai, et al., ``A survey for in-context learning,'' \emph{arXiv preprint arXiv:2301.00234}, 2023.

\end{thebibliography}

\end{document}